\documentclass[review]{elsarticle}

\usepackage{lineno,hyperref}
\usepackage{amsmath}
\usepackage{tikz}
\usetikzlibrary{positioning}
\modulolinenumbers[5]

\journal{JVCI}

\begin{document}

\begin{frontmatter}

\title{Generalization of feature embeddings transferred from different video anomaly detection domains}

\author{Fernando P. dos Santos}
\ead{fernando\_persan@usp.br}

\author{Leonardo S. F. Ribeiro}
\ead{leonardo.sampaio.ribeiro@usp.br}

\author{Moacir A. Ponti}
\address{Institute of Mathematical and Computer Sciences (ICMC), University of S\~ao Paulo (USP), S\~ao Carlos/SP, Brazil, 13566-590}
\ead[url]{http://www.icmc.usp.br/~moacir}
\ead{ponti@usp.br}

\begin{abstract}

Detecting anomalous activity in video surveillance often involves using only normal activity data in order to learn an accurate detector. Due to lack of annotated data for some specific target domain, one could employ existing data from a source domain to produce better predictions. Hence, transfer learning presents itself as an important tool. But how to analyze the resulting data space? This paper investigates video anomaly detection, in particular feature embeddings of pre-trained CNN that can be used with non-fully supervised data. By proposing novel cross-domain generalization measures, we study how source features can generalize for different target video domains, as well as analyze unsupervised transfer learning. The proposed generalization measures are not only a theorical approach, but show to be useful in practice as a way to understand which datasets can be used or transferred to describe video frames, which it is possible to better discriminate between normal and anomalous activity.

\end{abstract}

\end{frontmatter}


\section{Introduction}

An anomaly detection algorithm infers a model that is able to discriminate between a normal pattern and abnormal ones. In this context, \emph{learning} means inferring a function $f:X \rightarrow Y$ from a training set of examples $x_i \in X$, in which $X$ is an input space from a given domain, and it is composed of observations from a single class, which is referred to as normal, and $Y = \left\lbrace -1, +1 \right\rbrace$ is the output space, which in the case of anomaly detection refers to either ``normal'' or ``anomaly''~\cite{Chandola2009anomaly}. Most techniques focus on defining the normal activities in terms of a given problem, and therefore considering any event that deviates from such normality as anomalous. This approach is employed in surveillance of human crowds~\cite{Guo2016, Hu2016dense, Chaker2017}, pedestrian detection~\cite{Roshtkhari2013, Hasan2016, Ponti2017b}, and analysis of directions (human or vehicle motion)~\cite{Epaillard2016}. It is not however the only feasible view of the problem, and we believe that a good solution starts from recognizing the limitations of those systems: since only observations of normal activity or events are available for training, it is essential to have both sufficient data and an adequate input space to be able to create an accurate detector.

Real world problems become increasingly challenging when it comes to meeting the requirements of (annotated) data availability assumptions. First, the amount of available data collected from some task is often sufficient only to the same problem or domain~\cite{Aytar2011, Tommasi2010}. Second, many algorithms, in particular those with a large number of parameters to be learned such as deep learning methods, need a large amount of labeled data~\cite{Ravishankar2016}. However, annotating large amounts of data for learning can be expensive~\cite{Duan2012, li2015projected}. With these considerations, there is an immense incentive to investigate new techniques that can reduce the need for new labels and data~\cite{Long2015}. One set of techniques designed to that purpose is \textit{transfer learning}, first applied in the context of anomaly detection by Xu et. al~\cite{Xu2017}.

The goal of transfer learning is to supply a framework to solve new problems, using previously acquired knowledge from other similar solutions, quickly and effectively~\cite{Lu2015}. In this learning context, the first base (source) provides sufficient knowledge to recognize the desired information in the second base (target)~\cite{Yosinski2014}. Therefore, the main challenge in transfer learning is to correlate the distribution of training data from a source to the distribution of test data from a target~\cite{Hu2015}. However, if the source and target are similar in their domains, data with similar distributions, it is expected that the same classifier performs similarly across them~\cite{Tzeng2015}.

Transfer learning has been widely used in many different applications, such as image representation~\cite{Oquab2014}, facial attribute classification~\cite{Zhuang2018}, network traffic~\cite{Sun2018}, medical diagnosis~\cite{Cheng2018}, and pattern recognition~\cite{Wang2017}. In problems of video anomaly detection, both natural or urban scenarios, the assumption that similar problems have similar data distribution may not hold true due to factors such as variable illumination, camera perspective, and amount of clutter in the scene~\cite{Roshtkhari2013,Hao2017multi}. These factors contribute to the distancing between the activities and visual content of different domains. More importantly, the definition of anomaly itself may differ. Although in general an anomaly is an observation containing an abnormal pattern~\cite{Jiang2009}, this can be linked to different events as, for example, the appearance of a clandestine boat for the application of water surveillance, or a car parking on a pedestrian boardwalk in a traffic surveillance scenario. Consequently, detecting anomalies in video requires an arduous task of studying which attributes are relevant to well model the activities, allowing systems to distinguish normal from unusual ones. 

\begin{figure}[!ht]
\centering
\includegraphics[width=4in]{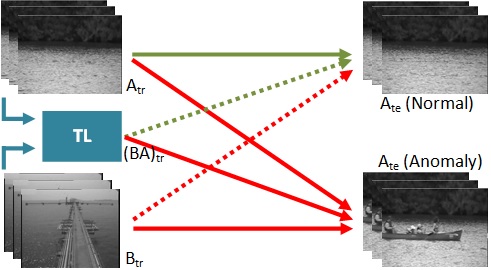}
\caption{Likely scenario in cross-domain application of anomaly detection. The green arrows represent normal events detection and red arrows identify anomalous occurrences from two distinct scenarios (sources $A$ and $B$); notice how normal events on the target domain are identified as anomalous due to the domain gap. It is expected that applying transfer learning methods (TL) between those two domains will significantly improve normal events detection on the target domain (represented by dashed flows).}
\label{fig:anomalyDetection}
\end{figure}

Consider the task of anomaly detection in surveillance videos and any two distinct domains $A$ and $B$: in this scenario, one can train a recognition system within a source domain $A$, in which $f:A \rightarrow Y$ may yield a reliable system for such domain, but could the same system be considered reliable even when used within the context of some target domain $B$? Intuition instructs that a likely scenario is that in which all samples from $B$ will be identified as anomalous (i.e. $f(b_i) = +1$ for all $b_i \in B$), therefore the system will fail at the same task. The challenge of transfer learning here is to design systems capable of performing \textit{consistently}, even if they were trained on samples from $A$ and used on samples from $B$. Fig.~\ref{fig:anomalyDetection} illustrates this scenario in which two distinct domains training sets ($A_{tr}$ and $B_{tr}$) are used to identify normal and anomalous activity of a single test domain ($A_{te}$). By using transfer learning techniques, a new source domain ($BA_{tr}$) is found, closing the gap between the data distributions of $A$ and $B$.

In this paper we investigate the \textit{generalization} of feature spaces extracted via a pre-trained Convolutional Neural Network (CNN) -- which does not require additional labels -- within the task of detecting anomalous activity in videos. CNNs have been successful in many computer vision tasks, being considered to be smart feature extraction modules that offer flexibility and a good level of cross-domain transfer learning~\cite{Lu2015, Razavian2014,  sun2017vehicle,Cavallari2018unsupervised}, even on video recognition tasks~\cite{Hu2016dense}. Investigating transfer learning in this scenario is relevant because CNN models are known to require large amounts of labeled training data in order to converge and to have any true learning guarantee; in practice many datasets do not have enough samples to allow training of deep networks from scratch. Leveraging pre-acquired knowledge is therefore a must and we find it important to highlight that recent research~\cite{MelloFerreiraPonti2017} has shown that theoretical learning guarantees are achieved when CNNs are trained with large datasets such as ImageNet~\cite{Russakovsky2015}.

We aim to understand the benefits of using CNN-based feature embeddings coupled with classic and modern transfer learning frameworks in the task of anomaly detection. We designed experiments on transferring an anomaly detector's knowledge within: (i) the cross-domain feature embeddings; (ii) cross-domain Principal Component Analysis (PCA)~\cite{Jolliffe1986}; and (iii) Transfer Component Analysis (TCA)~\cite{Pan2011}. More than designing a framework for transfer learning within the task of anomaly detection, we introduce a novel evaluation approach regarding generalization of feature embeddings between distinct domains and present the performance of our systems within these new metrics. Our contributions are then two-fold: (i) we developed a framework for transfer learning applied in the task of anomaly detection in videos; and (ii) we designed  a novel evaluation approach regarding generalization of feature embeddings. 

\section{Methodology}

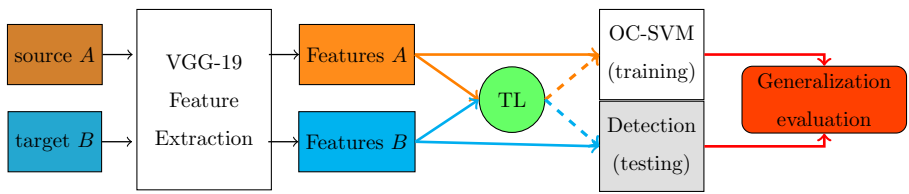
\begin{figure}[!ht]
\begin{center}
    \begin{tikzpicture}[node distance=0.45cm,scale=2, every node/.style={scale=0.8}]

    \node[draw, text centered, minimum size=0.75cm, text width=2.0cm, minimum height=3cm] (featext) {VGG-19 Feature Extraction};
    
    \node[draw,minimum size=1cm, name=input, left=of featext, yshift=0.75cm, fill=gray!35!orange] (inputA) {source $A$};
    \node[draw,minimum size=1cm, below of=inputA, yshift=-1cm, fill=gray!35!cyan] (inputB) {target $B$};    
    
    \node[draw,minimum size=1cm, right=of inputA, xshift=2.7cm, fill=orange!90!white] (codeA) {Features $A$};
    \node[draw,minimum size=1cm, right=of inputB, xshift=2.7cm, fill=cyan!90!white] (codeB) {Features $B$};
    
    \node[draw,circle,minimum size=0.75cm, right=of codeA, xshift=0.5cm, yshift=-0.75cm, fill=green!60!white, text width=0.75cm, align=center] (tl) {TL};
    
    \node[draw, minimum size=1cm, right=of codeA, xshift=2.5cm, minimum height=1.5cm, text width=1.5cm, text centered, fill=white] (train) {OC-SVM (training)};
    \node[draw, minimum size=1cm, minimum height=1.5cm, text width=1.5cm, text centered, below=of train, yshift=0.55cm, fill=gray!25!white] (test) {Detection (testing)};
    
    \node[draw,rounded corners, minimum size=1cm, minimum height=1.1cm, text width=2.5cm, text centered, right=of tl, xshift=2.7cm, fill=red!50!orange] (gener) {Generalization evaluation};
    
    \draw[->,line width=0.5pt] (inputA.east) -- ++(0.2,0);
    \draw[->,line width=0.5pt] (inputB.east) -- ++(0.2,0);
    \draw[<-,line width=0.5pt] (codeA.west) -- ++(-0.2,0);
    \draw[<-,line width=0.5pt] (codeB.west) -- ++(-0.2,0);
    \draw[->,line width=1.0pt,orange] (codeA.east) -- (train.west);
    \draw[->,line width=1.0pt, orange] (codeA.east) -- (tl.west);
    \draw[->,line width=1.2pt, dashed, orange] (tl.east) -- (train.west);
    \draw[->,line width=1.0pt, cyan] (codeB.east) -- (tl.west);
    \draw[->,line width=1.2pt, dashed, cyan] (tl.east) -- (test.west);
    \draw[->,line width=1.0pt, cyan] (codeB.east) -- (test.west);

    \draw[->,line width=1.0pt, red] (train.east) -| (gener.north);
    \draw[->,line width=1.0pt, red] (test.east) -| (gener.south);
    \end{tikzpicture}
  \end{center}
  \caption{Experimental setup: both source and target domains feature spaces embedding are independently computed via the same deep network model, then the source $A$ is used to train an One Class SVM, while target $B$ is tested on this trained model. Transfer learning (TL) can be used to transform such spaces before training/testing (indicated in dashed lines).}\label{fig:method}
\end{figure}

In order to evaluate generalization, we use an experimental setup that considers several different domains, each coming from different video anomaly detection datasets. Figure~\ref{fig:method} depicts this setup: all datasets are individually mapped to a feature space using the same pre-trained VGG-19~\cite{Simonyan2014} network. For every domain pair (source $A$, target $B$), the training set of $A$ is used to train an anomaly detector model (One Class Support Vector Machine (OC-SVM)~\cite{Chen2001}), which is then employed on the test set of $B$. Afterwards, we evaluate the generalization between $A$ and $B$ and the impact of transfer learning methods on this generalization.

Three scenarios were investigated: (i) without any data transformation; (ii) with PCA applied across domains; and (iii) with TCA. The next sections detail each step: the feature extraction (Section~\ref{ss.featureextraction}) and transfer learning (Section~\ref{ss.transferlearning}) methods, as well as the generalization evaluation metric (Section~\ref{ss.generalizationmetric}).

\subsection{Feature Extraction}
\label{ss.featureextraction}

Feature extraction is an important step that directly influences the result of classifiers, an inappropriate descriptor choice can significantly degrade performance and accuracy. With the great number of feature extraction methods available and the particularities of each database, finding the relevant descriptor can become a trial-and-error task. To lessen this hindrance, data-driven approaches to feature extraction can be employed, of which deep learning was particularly shown to produce suitable representations. One of the main advantages presented by methods of feature learning in relation to handcrafted extraction is the generalization of the feature space produced for data not seen within the same visual domain~\cite{Sarraf2016, Zhou2017}. Among current deep learning techniques, CNNs are widely used to compute feature space representations by sharing information and internal connections~\cite{Liang2015}.

The VGG-19 CNN~\cite{Simonyan2014}, composed by its 19 weight layers, is widely used in the context of pre-trained CNNs due to its simple architecture: a composition of convolutional layers with $3 \times 3$ sized filters. The motivation of this structure is that two sequential $3 \times 3$ filters have an effective receptive field of a $5 \times 5$ and, with the addition of more rectification layers, the decision function is more discriminative. This concept can be expanded to replace $7 \times 7$ filters with three filters $3 \times 3$~\cite{Ponti2017}. After a sequence of convolutional layers (generally composed of the $3 \times 3$ filters) and max-pooling, the top of VGG-19 is composed of fully connected layers which aim to provide probabilities for trained classes. Each layer provides a new descriptor that can be used to describe shapes and edges (lower layers) and texture and semantics (higher layers)~\cite{Razavian2014}, each one with a number of predefined attributes.

We use this architecture as a feature extractor, each frame of the selected domains is forward-passed through a network with its weights pre-trained using the ImageNet dataset~\cite{Russakovsky2015} (and used as-is). To ensure compatibility with the pre-training phase, each frame has its resolution reduced to $224 \times 224$ and the last network layer is discarded due to its tie with the ImageNet classification task; the feature set is then a result of the second-to-last layer (FC-4096), a 4096-dimensional feature vector.

\subsection{Transfer Learning}
\label{ss.transferlearning}

For our baseline, the feature space obtained in the second-to-last layer in VGG-19 is applied in two circumstances: without pre-processing and with dimensionality reduction. Considering two domains ($A$ and $B$) and their respective training set ($A_{tr}$ and $B_{tr}$) and test set ($A_{te}$ and $B_{te}$), the cross-feature occurs with direct inversion of test sets in relation to their source domain. Therefore, the anomaly detection performed in $A_{te}$ uses the information contained in $B_{tr}$ and $A_{tr}$ is the basis for $B_{te}$. With this setup and without pre-processing, all $4096$ attributes available by the VGG-19 layer are considered in the evaluation of \textbf{scenario (i)}.

For \textbf{scenario (ii)}, we investigate how the classic space projection technique PCA~\cite{Jolliffe1986} may contribute to transfer learning. We apply PCA on the original feature space ($A_{tr}$) and learn a projection matrix $\Theta$ of this space to a subset of its eigenvectors. We experiment then using the same projection $\Theta$ on the feature space of another domain ($B_{te}$) and evaluate how well does learning trained over $A$ perform on $B$ given both spaces were projected over $A$'s eigenvectors.

Finally, we designed the \textbf{third scenario} where, to the best of our knowledge, TCA is applied for the first time to the task of anomaly detection over distinct domains. TCA, introduced by Pan et al.~\cite{Pan2011}, is motivated by the assumption that common factors exist between different domains. The goal of performing TCA is to project both feature spaces into a new, common space where the distance between samples from distinct domains is small and data variance is kept large (this latter objective being the same as designed by classic PCA). The solution showed by Pan et. al~\cite{Pan2011} is to formulate TCA as a variant of Kernelized-PCA, where data is centralized over both domains, which the Gram Matrix is defined as the composition of gram matrices, where the subscripts indicate the source ($S$) or target ($T$) domains and $K_{S, S} = X_S X_S^t$: 

\begin{equation}
    K = \begin{bmatrix} 
    K_{S, S}       & K_{S, T}  \\
    K_{T, S}       & K_{T, T} \\
\end{bmatrix}
\end{equation}

To build this matrix, TCA needs at least some training data from the second domain and $K$ is weighted to highlight intra-domain and diminish inter-domain dissimilarities, stimulating the eigen decomposition to better capture existing variance across-domains instead of in each individually. Using this approach, TCA is designed to find eigenvectors and eigenvalues in a combined feature space, therefore acting upon the assumption that common factors do exist between domains and finding a linear transformation into a space that highlights those common factors.

\subsection{A Generalization Metric for Cross-Domain Feature Spaces}
\label{ss.generalizationmetric}

The field of machine learning has long dealt with the the idea of developing theoretical guarantees and support for what is called ``learning'' within the context of each algorithm; by far the most stable theory comes in the form of Statistical Learning Theory (STL)~\cite{Vapnik1999overview, vonLuxburg2011}. STL has since its introduction been widely used to assess the quality of studies within machine learning field and, more specifically, to support the mathematical proofs that guarantee Support Vector Machines (SVM) generate optimum classifiers. One recent work in this direction explores how each guarantee can be applied for deep learning frameworks~\cite{MelloFerreiraPonti2017}. 

As one of the major contributions of this study, we hereafter propose a new metric to evaluate cross-domain transfer learning systems using tools provided by the SLT. Inspired on the evaluation of supervised learning models, which proved invaluable to researchers working to design such systems, we aim then to contribute to the field of cross-domain transfer learning by asking: how can one measure generalization of a feature space produced by some method? 
One of the main concepts that drives STL is generalization. In supervised learning, generalization is a divergence that measures how well a classifier performance with unseen data is \textit{consistent} with its performance on training (seen) data. It can be mathematically expressed as:
\begin{equation}
    |R_{emp}(f_n) - R(f_n)|,
\end{equation}
where $R(f_n)$ is the true risk (expectancy of loss) of a classifier $f_n$ over ``all data'', also called expected risk, and $R_{emp}(f_n)$ is the risk of the same $f_n$, but evaluated over the training set (the empirical sample), called empirical risk. The idea of true risk is purposefully abstract (being an intractable quantity), but it nonetheless serves its goal of highlighting the importance of not losing ourselves amid metrics of accuracy and cost over training data, metrics that may not paint the bigger picture of how well a system works.

With a similar state of STL, we believe that evaluation of transfer learning systems applied to anomaly detection cannot rely solely on classic metrics such as Receiver Operating Characteristic (ROC), and derived metrics such as Area Under the ROC Curve (AUC) and Equal Error Rate (EER); if we aim to measure how well a system trained within one domain's feature space performs on a dissimilar domain's space, the idea of generalization presents itself as a great fit. We propose then to adapt the idea in the form of two metrics: (i) Partial Cross-domain Feature Space Generalization (Partial CDFG, or $G_{part}$); and (ii) Complete Cross-domain Feature Space Generalization (Complete CDFG, or $G_{comp}$), defined as:

\begin{equation}
    G_{part}(f_n^A) = |\underset{x \in \mathcal{X}^A}{R(f_n^A)} - \underset{x \in \mathcal{X}^B}{R(f_n^A)}|
\end{equation}

\begin{equation}
    G_{comp}(f_n^A, f_n^B) = \dfrac{1}{2}\left(|\underset{x \in \mathcal{X}^A}{R(f_n^A)} - \underset{x \in \mathcal{X}^B}{R(f_n^A)}| + |\underset{x \in \mathcal{X}^B}{R(f_n^B)} - \underset{x \in \mathcal{X}^A}{R(f_n^B)}|\right)
\end{equation}
where $f_n^A$ and $f_n^B$ are classifiers found by a classification algorithm by the way of Empirical Risk Minimization applied over, respectively, domain $A$ and domain $B$; the expression $\underset{x \in \mathcal{X}^A}{R(f_n^A)}$ denotes the risk of classifier $f_n^A$ over the feature space $\mathcal{X}^A$ and $\underset{x \in \mathcal{X}^B}{R(f_n^A)}$ denotes the risk of the same classifier $f_n^A$ over the feature space of the second domain, $\mathcal{X}^B$; the mirrored definition being valid for $\underset{x \in \mathcal{X}^B}{R(f_n^B)}$ and $\underset{x \in \mathcal{X}^A}{R(f_n^B)}$. Even though the real risk is intractable, it could be approximated by test sets (data that is specially separated to represent unseen data) from each domain. Finally, transferring those ideas to the context of anomaly detection, we could define the empirical risk in this formulation as the evaluation of either $AUC$ or $EER$, both classic metrics for anomaly detection tasks.

Beside the formulation of the metrics, we also aim to guarantee their significance and meaningfulness by restricting the scenarios upon which it could be applied. Given two domains ($A$ and $B$) and their respective feature spaces ($\mathcal{X}^A$ and $\mathcal{X}^B$), $G_{part}$ and $G_{comp}$ are meaningful metrics if: 

\begin{itemize}
    \item The bias (set of admissible functions) of the classification/detection algorithms are the same (e.g. same parameters on an SVM training setup);
    \item The spaces $\mathcal{X}^A$ and $\mathcal{X}^B$ are composed of the same set of descriptors for both domains;
    \item The transfer learning or domain mapping method should have no prior knowledge of test data on either domain.
\end{itemize}



Evaluating with CDFG affords us a multi-leveled analysis of feature spaces and transfer learning systems. $G_{part}$ is a good representation of how well training over a domain $A$ is well-suited or adapted to work over samples from domain $B$ with consistent performance. It is however ''one-way'' regarding domain, i.e we are looking at the mapping or transferring characteristic only in the $A \longrightarrow B$ direction. It is, regardless of this limitation, relevant to the analysis, specially when used in conjunction with $G_{comp}$. While $G_{part}$ is partial to the chosen descriptors and how each domain is particularly well represented by such descriptors, $G_{comp}$ is a better measure of the quality of the transfer system itself and its robustness when tested over different contexts and pairs of domains.

We are going to hereafter introduce three particular analysis levels afforded by our novel metrics. To compare methodologies by pairs (defined by classification algorithm and transfer learning technique, and indicated by the subscripts $\alpha$ and $\beta$), we selected classifiers obtained through the principle of empirical risk minimization for each methodology, denoted as $f_\alpha$ and $f_\beta$. One can claim that a method is more \textit{generalizable} in different levels by satisfying inequalities, one of such expressed bellow:

\begin{equation}    
\label{eq:OneDirectionA}
G_{part}(f_\alpha^A) < G_{part}(f_\beta^A)
\end{equation}

With this relationship satisfied, one could claim that method $\alpha$ is capable of generalizing well from domain $A$ to domain $B$. One can also verify the $G_{part}$ metric from the ``opposite direction'' and assess if the $\alpha$ methodology is also better than $\beta$ at generalizing from $B$ to $A$, as expressed by the inequality:

\begin{equation} 
\label{eq:OneDirectionB}
G_{part}(f_\alpha^B) < G_{part}(f_\beta^B)
\end{equation}

Hence, evaluating the $G_{part}$ on both directions gives us an understanding of the first level of generalization: how well the space obtained from one domain is applicable to another and how this applicability is captured by the chosen methodology. These comparisons do not assess directly the transfer method, being influenced and capturing well aspects regarding the representability of each feature space. To obtain a more precise and rigorous analysis of the transfer learning method itself, we should compare using the $G_{comp}$ metric:

\begin{equation} 
\label{eq:TwoDirections}
G_{comp}(f_\alpha^A, f_\alpha^B) < G_{comp}(f_\beta^A, f_\beta^B)
\end{equation}

However, the best use of our metrics come from applying each $G_{part}$ and $G_{comp}$ at the same time, given that the $G_{comp}$ can be influenced by high discrepancy between the two $G_{part}$ that compose it. It is primal then that all three comparisons are taken into account in the assessment of any two competing methods.

\section{Experiments}

We recall the experimental setup depicted in Figure~\ref{fig:method}. First, we evaluate the feature spaces with anomaly detection. With these results, we apply our generalization metrics to compare the performance of domains and methods used. In the following sections we describe the datasets and detail the experiments.

\subsection{Datasets}

For transfer learning and feature generalization experiments, we use seven anomaly detection videos/datasets (natural and urban scenarios) differing in several aspects, including frame resolution, amount of training frames, illumination conditions, perspectives, and presence of clutter.

\underline{Natural scenarios} (water surveillance activities):
\begin{itemize}
    \item \textbf{Canoe}: a video with 1050 frames of $240 \times 320$ pixels, which 200 are separated for training. It has scenery of nature with a river in the center and a canoe which invades the waters indicating the anomaly~\cite{Jodoin2008};
    \item \textbf{Boat-River}: similar to Canoe, but with higher resolution ($576 \times 740$ pixels) and different perspective. It has only 80 training frames and the majority of frames with anomalies in a single video~\cite{Zaharescu2010};
    \item \textbf{Boat-Sea}: similar to Canoe and Boat-River, but with occlusion in the scenes, making it difficult to correlate with the first two videos. Boat-Sea is composed of a single video of $576 \times 720$ resolution, in which 100 first frames are for training and the remaining ones for tests~\cite{Zaharescu2010}. 
\end{itemize}

\underline{Urban scenarios} (traffic and pedestrian activities):
\begin{itemize}
    \item \textbf{UCSD Pedestrian 1 (Ped1)}: it consists of videos of footpath, in which the presence of pedestrians are considered normal, and abnormal activity includes cyclists, skaters, and others. UCSD Ped1 contains 34 videos for training and 36 videos for tests with $158 \times 238$ pixels of resolution~\cite{Mahadevan2010}.
    \item \textbf{UCSD Pedestrian 2 (Ped2)}: similar to Ped1, UCSD Ped2 contains 16 videos for training and 12 videos for tests with $240 \times 360$ pixels of resolution. However, the placement of the cameras is different between the datasets~\cite{Mahadevan2010}.
    \item \textbf{Belleview}: it has frames with $240 \times 320$ pixels in a single video, with images of a road intersection in which anomalies are characterized by vehicle conversions and the standard behavior by the straight-line pass~\cite{Zaharescu2010};
    \item \textbf{Train}: it is also a single video (the first 800 frames of $386 \times 288$ are used in training) where the illumination varies rapidly due to the passage of the train through the tunnels. The anomalies are composed by the movement of the people inside the wagons~\cite{Zaharescu2010}.
\end{itemize}

\begin{figure}[!ht]
\centering
\includegraphics[width=4.5in]{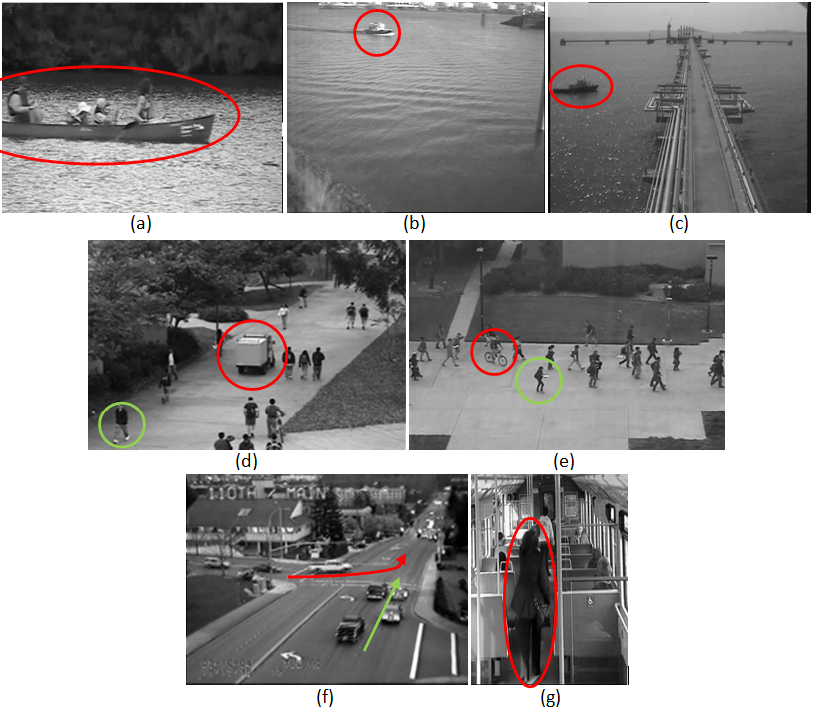}
\caption{Samples of test frames from: (a) Canoe; (b) Boat-River; (c) Boat-Sea; (d) UCSD-Ped1; (e) UCSD-Ped2; (f) Belleview; and (g) Train. Anomalous events are represent in red (boats, trucks, cyclists, vehicle conversions, and passenger movement). Examples of normal events are in green (pedestrians and straight-line pass).}
\label{fig:frameExamples}
\end{figure}

All frames from Canoe, Boat-River, Boat-Sea, and Train were converted to grayscale via: \textit{$0.299 R + 0.587 G + 0.114 B$}. The other urban datasets are originally in grayscale. Examples from all videos/datasets are presented in Fig.~\ref{fig:frameExamples}. Since the urban datasets are more complex and need to be described via motion/direction attributes, we expect the feature embeddings and transfer learning methods to work better within the Natural scenarios.

\subsection{Anomaly Detection, Parameters and Evaluation}

The Area Under the Curve (AUC) and Equal Error Rate (EER) are used to evaluate the anomaly detection in frame level. That means the model must predict which frames contain at least one anomaly and compare them with their respective ground-truth to determine the false positive rate (FPR) and true positive rate (TPR). The EER is the value on the ROC Curve in which $FPR = 1 - TPR$. As an anomaly detector, we used OC-SVM~\cite{Chen2001} that estimates support vectors having only positive labels in training set by including regularization to single out outliers. When the linear kernel is used, a hyper-sphere is drawn to cover the training samples, allowing an amount of outliers $v$ in $(0, 1]$. In our approach, we used $v = 0.25$.

For our generalization measure to be validated, we performed three experiments: first using the original feature embedding (Full VGG-19) with 4096 features; second, by applying PCA to these spaces selecting the 80 principal components (this value was chosen since Boat-River video has the smallest training set, 80 training examples, limiting the PCA analysis); third, the same dimensionality (80) is used in TCA method using the RBF kernel. In the original VGG-19, only the source training set is used to infer the model, that is then used to identify abnormal events using the target test set. For PCA, the transformation computed for the source training set is used to transform and select features for the target test set. In order to apply TCA, the matrix merged with the two training sets (source and target) is used to compute the transformation matrix, which is then used to select the principal components in target test set. With the results of the metrics (AUC and EER) the generalization measure is then computed. In that sense, we evaluate from a lower to a higher level of knowledge transfer. Table~\ref{table:TBPcaVariance} shows the PCA variance obtained with the videos/datasets: values near 1.0 in all sets indicate that there was no loss of relevant data in the created space with 80 features, due to the similarity among. 

\begin{table}[!ht]
\caption{PCA Variance with 80 principal components}
\label{table:TBPcaVariance}
\centering
\begin{tabular}{c|c}
\hline
\hline 
\textbf{Dataset} & \textbf{Variance}\\
\hline
\hline 
Canoe &	0.9996\\
Boat-River & 1.0\\
Boat-Sea & 0.9999\\
\hline
UCSD-Ped1 & 0.9992\\
UCSD-Ped2 & 0.9998\\
Belleview &	0.9997\\
Train & 0.9998\\
\hline 
\hline 
\end{tabular}
\end{table}

\section{Results and Discussion}

Table \ref{table:TBAnomalyNatural} presents the anomaly detection results of three experiments for natural scenarios: using the original VGG-19 feature embedding (Full VGG-19), after transformation with PCA (reduction to 80 dimensions), and applying unsupervised transfer learning by TCA (also with 80 dimensions). Considering the AUC and EER as performance metrics in the comparison of TCA with the other two approaches, it is observed that the TCA is better in 6 pairs (66.6\% of total) tested, mainly when the source domain is Boat-River. In the natural scenarios, the average AUC across all TCA sets was 86.68\%, meanwhile for Full VGG-19 was 70.47\%. As expected, the PCA space performance is outperformed by TCA, both in the number of pairs (1 vs 6), in the average AUC (69.52\% vs 86.88\%), and average EER (33.3\% vs 16.86\%). Additionally, there are expressive results of TCA in relation to Full VGG-19 and PCA: Boat-Sea $\longrightarrow$ Boat-River has an improvement (30.35\% to Full VGG-19) and Boat-River $\longrightarrow$ Boat-Sea in 31.41\% in relation to PCA. Therefore, as expected, the space provided by TCA overcomes Full VGG-19 and PCA in natural scenarios.

\begin{table}[!ht]
\caption{Anomaly Detection in natural scenarios (\%)}
\label{table:TBAnomalyNatural}
\centering
\begin{tabular}{c|cc|cc|cc}
\hline
\hline 
& \multicolumn{2}{c|}{\textbf{Full VGG-19}} & \multicolumn{2}{|c|}{\textbf{PCA}} & \multicolumn{2}{c}{\textbf{TCA}}\\
\textbf{Source $\longrightarrow$ Target} & \textbf{AUC} & \textbf{EER} & \textbf{AUC} & \textbf{EER} & \textbf{AUC} & \textbf{EER}\\
\hline 
\hline 
Canoe $\longrightarrow$ Canoe & \textbf{92.63} & \textbf{12.65} & 63.97 & 39.66 & 71.66 & 32.49\\
Boat-River $\longrightarrow$ Canoe & 92.75 & 12.65 & 53.61 & 48.1 & \textbf{99.1} & \textbf{3.04}\\
Boat-Sea $\longrightarrow$ Canoe & 92.75 & 12.65 & 85.44 & 18.56 & \textbf{97.5} & \textbf{4.64}\\
\hline 
Boat-River $\longrightarrow$ Boat-River & 63.24 & 36.75 & 74.35 & 25.64 & \textbf{90.59} & \textbf{9.4}\\
Canoe $\longrightarrow$ Boat-River & \textbf{63.24} & \textbf{36.75} & 50.42 & 49.57 & 61.11 & 38.88\\
Boat-Sea $\longrightarrow$ Boat-River & 64.52 & 35.47 & 61.96 & 38.03 & \textbf{94.87} & \textbf{5.12}\\
\hline
Boat-Sea $\longrightarrow$ Boat-Sea & 54.97 & 46.15 & \textbf{97.01} & \textbf{9.89} & 91.37 & 16.48\\
Canoe $\longrightarrow$ Boat-Sea & 55.0 & 46.15 & 83.39 & 26.37 & \textbf{86.99} & \textbf{19.79}\\
Boat-River $\longrightarrow$ Boat-Sea & 55.2 & 46.15 & 55.54 & 43.96 & \textbf{86.95} & \textbf{21.91}\\
\hline
\hline
Average & 70.47 & 28.37 & 69.52 & 33.3 & \textbf{86.88} & \textbf{16.86} \\
\hline 
\hline 
\end{tabular}
\end{table}

Also, considering the AUC and EER as performance metrics in urban scenarios (see Table~\ref{table:TBAnomalyUrban}), it is highlighted that Full VGG-19 is better in 7 pairs (16 pairs in total), mainly when the target domain is Ped2 or Belleview. However, the difference between the averages of Full VGG-19 and TCA is practically negligible: 63.84\% vs 62.8\% in AUC and 39.41\% vs 40.51\% in EER. Unlike the results with natural environments, urban domains with PCA presented positive results. It is important to emphasize that the concept of anomalies between these domains is very different, implying that the transfer learning should not be totally transferred (negative transfer). Hence, considering only domains with the same meaning of anomalies (Ped1 and Ped2), TCA stands out in relation to Full VGG-19 and PCA in averages of AUC and ERR.

\begin{table}[!ht]
\caption{Anomaly Detection in urban scenarios (\%)}
\label{table:TBAnomalyUrban}
\centering
\begin{tabular}{c|cc|cc|cc}
\hline
\hline 
& \multicolumn{2}{c|}{\textbf{Full VGG-19}} & \multicolumn{2}{c|}{\textbf{PCA}} & \multicolumn{2}{c}{\textbf{TCA}}\\
\textbf{Source $\longrightarrow$ Target} & \textbf{AUC} & \textbf{EER} & \textbf{AUC} & \textbf{EER} & \textbf{AUC} & \textbf{EER}\\
\hline 
\hline 
Ped1 $\longrightarrow$ Ped1 & 50.91 & 51.4 & \textbf{71.46} & \textbf{35.17} & 62.94 & 39.68\\
Ped2 $\longrightarrow$ Ped1 & 50.82 & 51.46 & \textbf{64.01} & \textbf{39.13} & 60.39 & 41.7\\
Belleview $\longrightarrow$ Ped1 & 51.77 & 50.75 & \textbf{76.12} & \textbf{30.56} & 58.86 & 45.89\\
Train $\longrightarrow$ Ped1 & 53.42 & 51.75 & 60.65 & 39.72 & \textbf{71.02} & \textbf{33.66}\\
\hline
Ped2 $\longrightarrow$ Ped2 & \textbf{80.34} & \textbf{26.26} & 55.24 & 44.13 & 74.16 & 33.26\\
Ped1 $\longrightarrow$ Ped2& \textbf{80.18} & \textbf{26.25} & 56.95 & 46.14 & 67.06 & 38.54\\
Belleview $\longrightarrow$ Ped2 & \textbf{80.88} & \textbf{26.26} & 69.46 & 34.63 & 65.16 & 38.01\\
Train $\longrightarrow$ Ped2 & \textbf{81.81} & \textbf{25.69} & 61.77 & 41.89 & 50.11 & 52.51\\
\hline
Belleview $\longrightarrow$ Belleview & 68.91 & 33.47 & 50.54 & 51.38 & \textbf{72.63} & \textbf{32.24}\\
Ped1 $\longrightarrow$ Belleview & \textbf{68.67} & \textbf{33.45} & 56.22 & 45.45 & 68.39 & 35.25\\
Ped2 $\longrightarrow$ Belleview & \textbf{68.73} & \textbf{33.47} & 60.42 & 40.63 & 65.24 & 39.12\\
Train $\longrightarrow$ Belleview & \textbf{69.1} & \textbf{32.92} & 54.36 & 49.31 & 68.65 & 34.35\\
\hline
Train $\longrightarrow$ Train & 53.97 & 47.2 & \textbf{57.67} & \textbf{42.84} & 51.88 & 51.47\\
Ped1 $\longrightarrow$ Train & 54.02 & 46.73 & \textbf{57.75} & \textbf{46.16} & 53.98 & 43.96\\
Ped2 $\longrightarrow$ Train & 54.13 & 46.67 & 55.47 & 49.0 & \textbf{55.56} & \textbf{42.68}\\
Belleview $\longrightarrow$ Train & 53.85 & 46.84 & 50.63 & 51.46 & \textbf{58.86} & \textbf{45.89}\\
\hline 
\hline 
Average & \textbf{63.84} & \textbf{39.41} & 59.92 & 42.97 & 62.8 & 40.51\\
\hline
\hline
\end{tabular}
\end{table}

\begin{figure}[!ht]
\centering
\begin{tabular}{c}
    \includegraphics[width=0.45\linewidth]{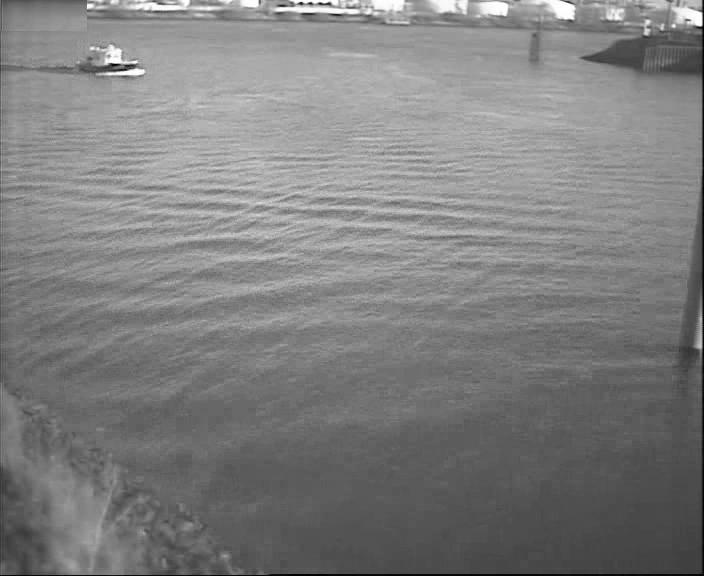} 
    \includegraphics[width=0.45\linewidth]{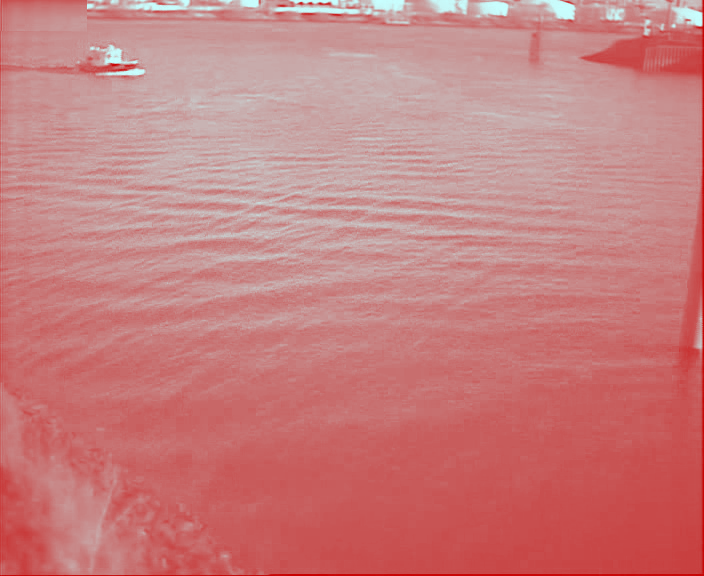} \\
\end{tabular}
\caption{Example of same frame from Boat-River considered normal with Full VGG-19 (left) and anomalous with TCA (right).}
\label{fig:RiverAnomaly}
\end{figure}

Although the transfer learning by TCA be superior to original space (Full VGG-19) and PCA, those metrics (AUC and EER) are not enough to guarantee the feature space generalization. Analyzing the results in isolation gives an imprecision due to the great variety of performances achieved. For these reasons, our generalization metrics offers a more detailed and reliable comparison if one methodology overcomes other. Evidently, generalization does not depend only on the techniques, but also on the similarity among domains. In Figure~\ref{fig:RiverAnomaly} is presented a frame of the Boat-River video, pointed out as anomalous, which erroneously was detected as "normal" using Full VGG-19 (Boat-Sea $\longrightarrow$ Boat-River). However, applying TCA the anomaly was detected (Boat-Sea $\longrightarrow$ Boat-River).

\subsection{Transferred Features Generalization}

Based on the metrics used to anomaly detection (AUC and EER) on generalization measure proposed in this paper, we evaluated the performance of the feature spaces provided by Full VGG-19 cross-domain, PCA cross-domain, and transfer learning by TCA. First, we evaluate the generalization at the first level, $G_{part}$. The results using inequations (\ref{eq:OneDirectionA}) and (\ref{eq:OneDirectionB}) are presented in Table~\ref{table:TBOneWay}. In general, the transfer learning method TCA is superior in the two metrics evaluated in the $G_{part}$. Considering only AUC, the average of all cases (12 sets) of TCA was 8.47\%, with Full VGG-19 in 22.43\% and PCA in 17.27\%. The same occurs with EER, being the best rate for TCA with 8.1\%. In terms of similarity between domains, Canoe and Boat-Sea are very close in the feature space mapping, in both directions TCA performs with high transfer rates. There is also great applicability of the feature spaces in contexts of different anomalies, Belleview and Ped1, in which the transfer learning is more significant from Ped1 to Belleview. It is also observed that Ped1 offers high learning rates for Ped2, however the inverse does not occur in the same intensity. Another highlight is the higher performance of the PCA when compared to Full VGG-19, demonstrating that the dimensionality reduction increases the performance during cross-feature. These data contradict the isolated analysis from Tables~\ref{table:TBAnomalyNatural} and~\ref{table:TBAnomalyUrban}, implicitly the importance of our generalization metrics.

\begin{table}[!ht]
\caption{Partial Cross-domain Feature Space Generalization ($G_{part}$) (\%)}
\label{table:TBOneWay}
\centering
\begin{tabular}{c|cc|cc|cc}
\hline
\hline 
& \multicolumn{2}{c|}{\textbf{Full VGG-19}} & \multicolumn{2}{c|}{\textbf{PCA}} & \multicolumn{2}{c}{\textbf{TCA}}\\
\textbf{Source $\longrightarrow$ Target} & \textbf{AUC} & \textbf{EER} & \textbf{AUC} & \textbf{EER} & \textbf{AUC} & \textbf{EER}\\
\hline 
\hline 
Boat-River $\longrightarrow$ Canoe & 29.51 & 24.1 & 20.74 & 22.46 & \textbf{8.51} & \textbf{6.36}\\
Boat-Sea $\longrightarrow$ Canoe & 37.78 & 33.5 & 11.57 & 8.67 & \textbf{6.13} & \textbf{11.84}\\
\hline
Canoe $\longrightarrow$ Boat-River & 29.39 & 24.1 & 13.55 & 9.91 & \textbf{10.55} & \textbf{6.39}\\
Boat-Sea $\longrightarrow$ Boat-River	& 9.55 & 10.68 & 35.05 & 28.14 & \textbf{3.5} & \textbf{11.36}\\
\hline
Canoe $\longrightarrow$ Boat-Sea & 37.63 & 33.5 & 19.42 & 13.29 & \textbf{15.33} & \textbf{12.7}\\
Boat-River $\longrightarrow$ Boat-Sea & 8.04 & 9.4 & 18.81 & 18.32 & \textbf{3.64} & \textbf{12.51}\\
\hline 
\hline
Ped2 $\longrightarrow$ Ped1 & 29.52 & 25.2 & \textbf{8.77} & \textbf{5.0} & 13.77 & 8.44\\
Belleview $\longrightarrow$ Ped1 & 17.14 & 17.28 & 25.58 & 20.82 & \textbf{14.27} & \textbf{10.47}\\
\hline
Ped1 $\longrightarrow$ Ped2 & 29.27 & 25.15 & 14.51 & 10.97 & \textbf{4.12} & \textbf{1.14}\\
Belleview $\longrightarrow$ Ped2 & 11.97 & 7.21	& 18.92 & 16.75 & \textbf{7.47} & \textbf{5.77}\\
\hline
Ped1 $\longrightarrow$ Belleview & 17.76 & 17.95 & 15.24 & 10.28 & \textbf{5.45} & \textbf{4.43}\\
Ped2 $\longrightarrow$ Belleview & 11.61 & 7.21	& \textbf{5.18} & \textbf{3.5} & 8.92 & 5.86\\
\hline 
\hline 
Average & 22.43 & 19.6 & 17.27 & 14.0 & \textbf{8.47} & \textbf{8.10}\\
\hline 
\hline 
\end{tabular}
\end{table}

The generalization in the first level excludes more complex and pertinent aspects to the generated spaces. The $G_{part}$ analysis is not enough to verify the immersion of two domains for a unique model, except in cases where there will be only contribution from one domain to other, without the need of the inversion from source and target. This scenario is noticeable in situations where the source is composed of large amounts of data and, therefore, it is sufficient to provide information to itself, not requiring auxiliary domains or prior learning. In more complex and accurate scenarios, $G_{comp}$ offers a deeper analysis of the proposed transfer learning model. In this approach, results in Table~\ref{table:TBTwoWays}, TCA offers even more generalization in relation to other two methods. Among similar domains, the latent space created by TCA is highly applicable: Boat-River and Boat-Sea with 3.57\%; Boat-River and Canoe with 9.53\%; and Ped1 and Ped2 with 8.95\%. Even in domains with different anomalies, the performance gain is evidenced (Ped2 and Belleview with 8.19\%).

\begin{table}[!ht]
\caption{Complete Cross-domain Feature Space Generalization ($G_{comp}$) (\%)}
\label{table:TBTwoWays}
\centering
\begin{tabular}{c|cc|cc|cc}
\hline
\hline 
& \multicolumn{2}{c|}{\textbf{Full VGG-19}} & \multicolumn{2}{c|}{\textbf{PCA}} & \multicolumn{2}{c}{\textbf{TCA}}\\
\textbf{Datasets} & \textbf{AUC} & \textbf{EER} & \textbf{AUC} & \textbf{EER} & \textbf{AUC} & \textbf{EER}\\
\hline 
\hline 
(Canoe, Boat-River) & 29.45 & 24.1 & 17.14 & 16.18 & \textbf{9.53} & \textbf{6.37}\\
(Canoe, Boat-Sea) & 37.70 & 33.5 & 15.49 & 10.98 & \textbf{10.73} & \textbf{12.27}\\
(Boat-River, Boat-Sea) & 8.79 & 10.04 & 26.93 & 23.23 & \textbf{3.57} & \textbf{11.93}\\
\hline
(Ped1, Ped2) & 29.4 & 25.2 & 11.6 & 7.99 & \textbf{8.95} & \textbf{4.79}\\
(Ped1, Belleview) & 17.5 & 17.6 & 20.4 & 15.6 & \textbf{9.86} & \textbf{7.45}\\
(Ped2, Belleview) &	11.79 & 7.21 & 12.05 & 10.12 & \textbf{8.19} & \textbf{5.82}\\
\hline 
\hline 
\end{tabular}
\end{table}

For a full feature generalization, the three inequations (\ref{eq:OneDirectionA}, \ref{eq:OneDirectionB} and \ref{eq:TwoDirections}) must be satisfied. This level guarantees that the $G_{comp}$ is  contemplated without one $G_{part}$ compensating the other. In our experiments, it is observed that there is a compensation in two approved cases of $G_{comp}$ with TCA: (Ped1, Ped2) and (Belleview, Ped2). Although Ped1 and Ped2 have the same concept of anomalies, the position of the cameras hinders the direct transfer learning, requiring preprocessing methods to facilitate the use of previously acquired knowledge. Despite the urban scenario, the concept of anomalies between Belleview and Ped2 is  different both semantically and visually: Belleview targets vehicles conversion, while Ped2 anomalies are related to the presence of vehicles on the scene.

\subsection{Negative Transfer}

A major concern in transfer learning methods is to apply only the acquired knowledge that favors the improvement of the task for the new target domain. For this purpose it is important to evaluate if the source domain is sufficiently related to target domain so that the transfer does not fail, causing the negative transfer~\cite{torrey2010transfer}. The negative transfer is evidenced when the transfer learning method achieved a lower performance to a method that does not performs transfer learning~\cite{pan2010survey}. In this context, $G_{part}$ and $G_{comp}$ should be applied to measure if the source domain or the methodology are suitable for a designated task.

\begin{table}[!ht]
\caption{Partial Cross-domain Feature Space Generalization ($G_{part}$)(\%): Negative Transfer}
\label{table:TBOneWayNegative}
\centering
\begin{tabular}{c|cc|cc|cc}
\hline
\hline 
& \multicolumn{2}{c|}{\textbf{Full VGG-19}} & \multicolumn{2}{c|}{\textbf{PCA}} & \multicolumn{2}{c}{\textbf{TCA}}\\
\textbf{Source $\longrightarrow$ Target} & \textbf{AUC} & \textbf{EER} & \textbf{AUC} & \textbf{EER} & \textbf{AUC} & \textbf{EER}\\
\hline 
\hline 
Train $\longrightarrow$ Ped1 & \textbf{0.55} & \textbf{4.55} & 2.98 & 3.12 & 19.14 & 17.81\\
Train $\longrightarrow$ Ped2 & 27.84 & 21.51 & 4.1 & 0.95 & \textbf{1.77} & \textbf{1.04}\\
Train $\longrightarrow$ Belleview & 15.13 & 14.28 & \textbf{3.31} & \textbf{6.47} & 16.77 & 17.12\\
\hline
Ped1 $\longrightarrow$ Train & \textbf{3.11} & \textbf{4.67} & 13.71 & 10.99 & 8.96 & 4.28\\
Ped2 $\longrightarrow$ Train & 26.21 & 20.41 & \textbf{0.23} & \textbf{4.87} & 18.6 & 9.42\\
Belleview $\longrightarrow$ Train & 15.06 & 13.37 & \textbf{0.09} & \textbf{0.08} & 13.77 & 13.65\\
\hline
\hline
\end{tabular}
\end{table}


\begin{table}[!ht]
\caption{Complete Cross-domain Feature Space Generalization ($G_{comp}$)(\%): Negative Transfer}
\label{table:TBTwoWaysNegative}
\centering
\begin{tabular}{c|cc|cc|cc}
\hline
\hline 
& \multicolumn{2}{c|}{\textbf{Full VGG-19}} & \multicolumn{2}{c|}{\textbf{PCA}} & \multicolumn{2}{c}{\textbf{TCA}}\\
\textbf{Datasets} & \textbf{AUC} & \textbf{EER} & \textbf{AUC} & \textbf{EER} & \textbf{AUC} & \textbf{EER}\\
\hline 
\hline
(Train, Ped1) & \textbf{1.83} & \textbf{4.61} & 8.35 & 7.06 & 14.1 & 11.0\\
(Train, Ped2) & 27.0 & 21.0 & \textbf{2.17} & \textbf{2.91} & 10.2 & 5.23\\
(Train, Belleview) & 15.1 & 13.8 & \textbf{1.7} & \textbf{3.28} & 15.3 & 15.4\\
\hline
\hline
\end{tabular}
\end{table}


Tables~\ref{table:TBOneWayNegative} and~\ref{table:TBTwoWaysNegative} present correlation results between videos/datasets of urban scenarios, more specifically between Train and the others (Ped1, Ped2, and Belleview). Train presents concept of anomalies very different from the others, in which the dissimilarity between them (background and objects) are highly perceivable. By the analysis of $G_{part}$, it is observed that there is applicability of transfer learning only from Train $\longrightarrow$ Ped2. Differently, the other methods (Full VGG-19 and PCA) performed better than TCA, characterizing a negative transfer scenario. Consequently, $G_{comp}$ indicates that Train is not a suitable domain for Ped1, Ped2, or Belleview.

\subsection{A Closer Look on the Practical Application of the Generalization Metric}

The datasets studied in this paper have two levels of difficulty: the water surveillance videos are easier, depending mainly on the appearance of the objects on the frame, while the urban ones depend on more complex attributes such as motion and orientation (not explicitly captured via the employed VGG-19 features), and have a larger training set. In any case, the use of the full feature embedding shows similar results for a fixed target domain $B$, and any source domain $A$. As an example, for the Boat-River dataset, training an anomaly detector on any source domain, produces an EER of $\sim 36\%$. However, the generalization measure is not the same and indicates the best potential for transfer learning via TCA. 

\begin{figure}[!ht]
\centering
\includegraphics[width=1\linewidth]{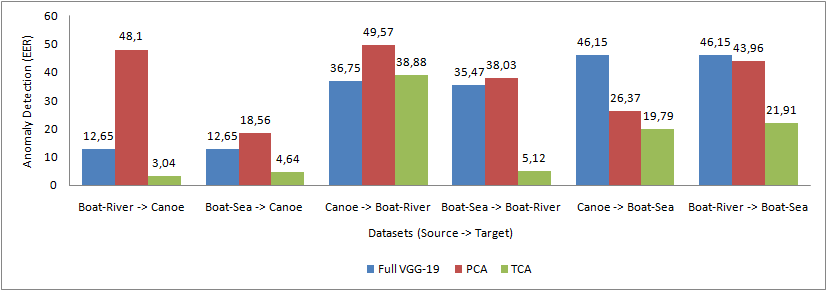}\\
\includegraphics[width=1\linewidth]{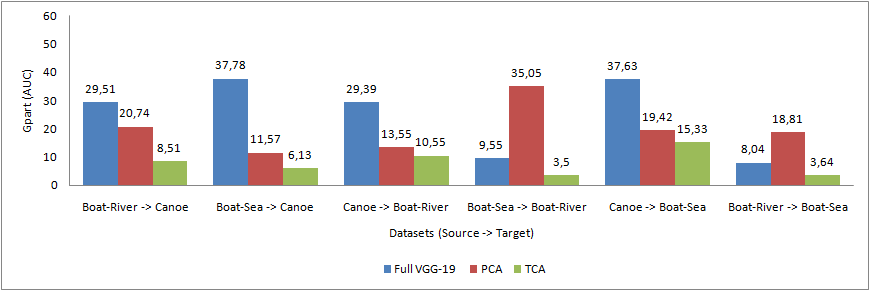}\\
\caption{Generalization from different water surveillance domains: (top) anomaly detection in EER; and (bottom) $G_{part}$ in AUC.}
\label{fig:targetWater}
\end{figure}

\begin{figure}[!ht]
\centering
\begin{tabular}{c}
\includegraphics[width=0.5\linewidth]{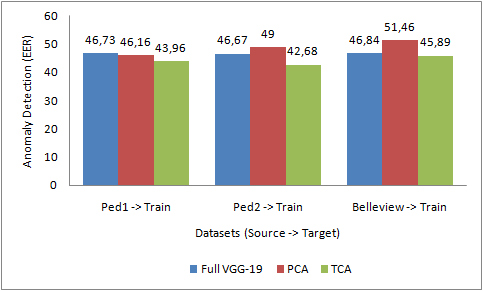}
\includegraphics[width=0.5\linewidth]{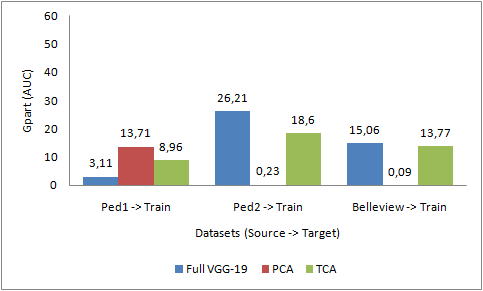}\\
\end{tabular}
\caption{Generalization from different source datasets to Train. The left barplots shows the EER (\%) from anomaly detection results, while the right barplots from the $G_\text{part}$ values.}
\label{fig:targetUrban}
\end{figure}

In Figure~\ref{fig:targetWater}, it is possible to see that, when evaluating the original feature embedding using both anomaly detection and the $G_{part}$, the dataset with the best (lower) generalization is a better potential as a source domain. If so, then the TCA method shows the best results. In the case of Boat-River (Figure~\ref{fig:targetWater}-top), the lowest EER are for Boat-Sea, which indeed, present the best result after application of TCA. Similar results are observed for Boat-Sea, for which the lowest $G_{part}$ is Boat-River, which also produces good improvement after applying TCA. For Train video as target, in Figure~\ref{fig:targetUrban}, the Ped1, Ped2, and Belleview were not possible to improve the results using transfer learning in the original feature embedding, which is clear when TCA often showed worse $G_{part}$ when compared to either PCA or the VGG features. 

\subsection{Running time analysis}

In our experiments, the frame per second processing rate was also analyzed. We consider two groups: water surveillance videos; and only UCSD datasets. The first one examines the average among all combinations of Canoe, Boat-River, and Boat-Sea. The second group considered the average when using datasets UCSD Ped1 and Ped2 only. Those experiments were grouped to evaluate scenarios considering videos with different duration.
After running all experiments, we discarded the highest and the lowest recorded times to avoid outliers. In Table~\ref{table:TBTime} we present the processing rates of frames per second (FPS), as well as the range of frames processed in the experiments for each group of videos.

All experiments were performed on the same machine, configured with an Intel(R) Core(TM) i7-7700K CPU, 64GB of RAM and a GeForce GTX 1080 Ti GPU. It is important to highlight that feature extraction using feed-forward through neural networks was performed using the GPU hardware while PCA, TCA and SVM computations were peformed on CPU.

\begin{table}[!ht]
\caption{Frame per seconds (FPS) using the two investigated approaches considering the stages: feature extraction (CNN), space transformation (PCA or TCA) and detection (OC-SVM)}
\label{table:TBTime}
\centering
\begin{tabular}{l|r|r|r}
\hline
\hline 
Video (\# of analyzed frames)&\textbf{Full VGG-19} & \textbf{PCA} & \textbf{TCA} \\
\hline 
\hline
Water (395 to 1162) & 1256 FPS & 2247 FPS & 1754 FPS  \\
UCSD (7110 to 20800) & 47 FPS & 1700 FPS & 4 FPS  \\
\hline
\hline
\end{tabular}
\end{table}

Note that, overall, PCA presents the highest FPS, since it only has to process the training samples from the source domain and, because it compacts the space, it allows for a faster OC-SVM detection. When more video frames are used for Transfer Learning (see UCSD row of Table~\ref{table:TBTime}), the computational running time performance of TCA is severely degraded due to the need of keeping a model with frames from both source and target videos. Again, PCA allows a fast projection into a lower dimensionality. In any case, even a rate of 4 FPS is not unfeasible considering our system was not optimized for detection, and that running TCA on GPU or using multiple cores of the CPU would significantly improve this performance.

\section{Conclusion}

A cross-domain generalization metric is able to complement evaluation of feature embeddings, indicating the potential for transfer learning. As our results indicate, when using CNN-based features, the TCA performance stands out and it is often accompanied by better generalization levels. This is a very interesting and simple approach that allows a guideline for the use of off-the-shelf feature extraction tools, boosting the performance of anomaly detection methods even when there is no additional data from the target domain that we are interested in solving.

We present experimental evidence that such generalization measures are not only theoretically, but can be useful in practice as a way to understand which datasets can be used as source or additional knowledge in order to describe video frames, hence it is possible to discriminate between normal and anomalous activity. This is important because it allows to use unsupervised or semi-supervised methods.

Transfer learning from video activity and other computer vision tasks is still a matter of future investigation. The proposed measures can be explored in the context of choosing which feature extraction method better suits some task, or to merge different datasets in order to accumulate a larger training set, and therefore increase learning guarantees.

\section*{Acknowledgment}

The authors would like to thank FAPESP for the grants \#2016/16111-4 and \#2017/22366-8, and CNPq (307973/2017-4). This study was financed in part by the Coordena\c{c}\~ao de Aperfei\c{c}oamento de Pessoal de N\'{i}vel Superior - Brasil (CAPES) - Finance Code 001. This work is also partially supported by the CEPID-CeMEAI (FAPESP grant \#2013/07375-0).

\section*{References}

\bibliography{mybibfile}

\end{document}